\documentclass[letterpaper, 10 pt, conference]{ieeeconf}  

\IEEEoverridecommandlockouts                              

\usepackage{float}  
\usepackage[font=scriptsize]{subfig} 
\usepackage{booktabs}  
\usepackage{makecell}  
\usepackage{array}
\usepackage{tabularx}

\usepackage{cite}
\usepackage{amsmath,amssymb,amsfonts}
\usepackage{algorithmic}
\usepackage{graphicx}  
\usepackage{textcomp}
\usepackage{xcolor}

\usepackage{hyperref}
\hypersetup{hypertex=true,colorlinks=true,linkcolor=blue,anchorcolor=blue,citecolor=blue}

\title{\LARGE \bf
A Click-Through Rate Prediction Method Based on Cross-Importance of Multi-Order Features
}

\author{Hao Wang$^{1}$ and Nao Li$^{*}$ 
	\\
	\textit{*Email: 2130101009@st.btbu.edu.cn}
	\thanks{*Nao Li is the corresponding author}
	\thanks{Hao Wang with School of Computer and Artificial Intelligence, Beijing Technology and Business University, Beijing, Haidian District, China
	}%
	\thanks{Nao Li with School of International Economics and Management, Beijing Technology and Business University,Beijing, China and State Key Laboratory of Resources and Environmental Information System, Institute of Geographic Sciences and Natural Resources Research, Chinese Academy of Sciences, Beijing, China 
	}
}

\begin{document}

\maketitle
\thispagestyle{empty}
\pagestyle{plain}  

\begin{abstract}
Most current click-through rate prediction(CTR) models create explicit or implicit high-order feature crosses through Hadamard product or inner product, with little attention to the importance of feature crossing; only few models are either limited to the second-order explicit feature crossing, implicitly to high-order feature crossing, or can learn the importance of high-order explicit feature crossing but fail to provide good interpretability for the model. This paper proposes a new model, FiiNet (Multiple Order Feature Interaction Importance Neural Networks). The model first uses the selective kernel network (SKNet) to explicitly construct multi-order feature crosses. It dynamically learns the importance of feature interaction combinations in a fine-grained manner, increasing the attention weight of important feature cross combinations and reducing the weight of featureless crosses. To verify that the FiiNet model can dynamically learn the importance of feature interaction combinations in a fine-grained manner and improve the model's recommendation performance and interpretability, this paper compares it with many click-through rate prediction models on two real datasets, proving that the FiiNet model incorporating the selective kernel network can effectively improve the recommendation effect and provide better interpretability. FiiNet model implementations are available in PyTorch\footnote{\url{https://github.com/whpython/FiiNet}}.
\end{abstract}


\section{Introduction}

Amidst rapid development of internet platforms and exponential growth of mobile end users, individuals now have access to a vast array of information in real-time through internet and terminal platforms. In the deluge of information, it becomes increasingly challenging for people to acquire information that is of interest and necessity to them. Amongst abundant data on videos, products, and reviews, it is difficult for the average person to effectively obtain useful information on a day-to-day basis\cite{b1}. The emergence of massive volumes of information contributes to a decrease in information utility rates, giving rise to the issue of information overload. Personalized recommendation technology, as a principal means of information filtering, can effectively alleviate the phenomenon of information overload and represents a core technology of user-oriented internet products. For instance, the University of Minnesota research group was the first to suggest the use of collaborative filtering, within their automated recommendation system GroupLens\cite{b2}, Amazon deployed collaborative filtering algorithms, marking the first recommendation system service to reach tens of millions of users and handle product scales in the millions\cite{b3}, The social platform Facebook employs machine learning algorithms for advertisement recommendation, inaugurating a new phase of feature engineering modeling and automation\cite{b4}, The streaming media platform YouTube implemented recommendations based on deep learning, which increased user stickiness\cite{b5}.\\
\indent Within the realm of recommendation tasks, the characteristics of datasets are generally high-dimensional and sparse. For example, if a user favors products from the Moutai brand and regularly purchases or browses coffee products (as coffee is of interest to the user), then it is highly probable that the user would like products resulting from their collaboration, such as the soy sauce-flavored latte. Likewise, if a user has an affinity for outdoor sports brand Under Armour's apparel and has recently been frequently watching NBA (National Basketball Association) games, and with the current season's player Stephen Curry being conferred the honor of FMVP (Finals Most Valuable Player), it indicates that the user may have a significant interest in joint-name jerseys and sportswear from the Under Armour brand in collaboration with Stephen Curry. Various features and their multi-order combinations play a crucial role in predicting user interests. Combinations between features, known as feature crosses, are instrumental in elevating certain performances of the recommendation system as evidenced by the aforementioned examples. 

\section{Related Work}

\subsection{Feature Crosses Model Related Work}

Typically, in business scenarios, the input to logistic regression (LR) models relies on manual design of feature crosses by professionals using prior knowledge. Factorization Machines (FM\cite{b6}) automatically design low-dimensional latent vectors for each feature in scene data, capturing the interactions of feature crosses through the training of these latent vectors. Field-aware Factorization Machines (FFM\cite{b7}) introduce the concept of feature fields, recognizing that different features belong to different fields. Attentional Factorization Machines (AFM\cite{b8}) employ an attention mechanism to learn the importance of different feature crosses within the model, etc. However, these methods are limited to exploring second-order feature crosses. Combining Deep Neural Networks (DNN) to implicitly construct high-order feature crosses is an important heuristic direction. Neural Factorization Machines (NFM\cite{b9}) design the bi-interaction layer in neural networks to multiply input vectors pairwise before feeding them into subsequent DNN layers for training. Wide \& Deep\cite{b5} combines the advantages of LR with deep neural networks (DNN) in a dual-tower structure, with LR module's manually designed feature crosses and DNN module's implicit high-order feature crosses. DeepFM\cite{b10} replaces the LR module in Wide \& Deep with an FM module, achieving complete automation in the model's design of feature crosses. Deep \& Cross\cite{b11} replaces the LR in Wide \& Deep with a special Cross network, automatically constructing high-order explicit feature crosses, allowing feature interactions to occur at the bit-wise level, and learning corresponding weights to avoid manual design of high-order feature crosses. xDeepFM\cite{b12}introduces the concept of vector-wise feature interaction within fields, transforming the cross into a CIN (compressed interaction network), which can explicitly undertake high-order crossing of feature embedding vectors, combined with DNN to implicitly construct high-order feature crosses. FiBiNET\cite{b13} employs a squeeze-and-excitation network layer (SENET\cite{b15}) within the channel attention mechanism to enhance feature embeddings, gaining information related to feature importance, and combining dot product and Hadamard product to perform feature crosses before being fed into subsequent DNN layers for training.AutoInt\cite{b14} constructs high-order feature crosses explicitly using a multi-head self-attention mechanism and can learn the importance of feature crosses.
\begin{table}[H]
	\caption{Describe the characteristics of feature crosses in different CTR models}
	\begin{center}
		\begin{tabular}{c|c|c|c}
			\toprule 
			\textbf{Model}&\textbf{type}&\textbf{order}&\textbf{importance} \\
			\midrule 
			LR&explicit&1& \\
			\hline
			FM\cite{b6}&explicit&2& \\
			FFM\cite{b7}&explicit&2& \\
			AFM\cite{b8}&explicit&2&$\checkmark$ \\
			\hline   
			Wide\&Deep\cite{b5}&implicit&3+& \\
			DeepFM\cite{b10}&implicit&3+& \\
			NFM\cite{b9}&implicit&3+& \\
			Deep\&Cross\cite{b11}&explicit&3+& \\
			xDeepFM\cite{b12}&explicit&3+& \\
			FiBiNET\cite{b13}&implicit&3+&$\checkmark$\\
			AutoInt\cite{b14}&explicit&3+&$\checkmark$\\
			FiiNet(\textbf{ours})&explicit&3+&$\checkmark$\\
			\bottomrule 
		\end{tabular}
		\label{tab1}
	\end{center}
\end{table}
Most current work creates explicit or implicit high-order feature crosses through Hadamard product or inner product, seldom focusing on the importance of feature crossing\cite{b13}. Here are some existing works that have paid attention to the importance of different feature crosses: AFM\cite{b8} uses an attention network to learn the importance of different feature crosses, but the model is limited to second-order explicit feature crosses. FiBiNET\cite{b13} uses the squeeze-and-excitation network layer (SENET\cite{b15}) in the channel attention mechanism to learn information about feature importance, combined with Hadamard product and dot product to create feature crosses, then the vectors are fed into the DNN layer to create implicit high-order feature crosses. AutoInt\cite{b14} uses a multi-head self-attention mechanism to explicitly construct high-order feature crosses and also learn the importance of different feature crosses, yet it cannot provide the model with good interpretability and recommendation effects. The above Table \ref{tab1} provides explanations on the importance of feature crosses in different click-through rate prediction models.

\subsection{Selective kernel network method}

SENet\cite{b15} explicitly models the interdependencies between convolutional feature channels to enhance the network's representational power in various image classification tasks and won first place in the ILSVRC 2017 classification challenge, demonstrating its success in image classification tasks. SKNet\cite{b16} introduces a dynamic selection mechanism within CNNs that allows each neuron to adjust the size of its receptive fields adaptively based on the input information's multiple scales. It is comprised of building blocks called Selective Kernel (SK) units, with multiple SK units stacked to form a deep Selective Kernel Network. In image classification benchmarks, like ImageNet and CIFAR, SKNet outperforms existing architectures such as SENet and ResNet with lower model complexity. In addition to image classification, there are other applications of SKNets. Using Selective Kernel Networks with attention can assist in early MRI-based diagnosis of Alzheimer's Disease\cite{b17}. Selective Kernel mechanisms have been helpful in improving accuracy in the field of remote sensing target detection\cite{b18}. An enhanced version of Selective Kernel Networks applying both spatial and channel attention, SKv2, has extended SKNet modules and achieved higher accuracy in image classification tasks\cite{b19}.\\
\indent It is widely recognized that different features and their combinations through feature crosses hold varying degrees of importance for recommendation tasks. For example, assume the target task is to predict an individual's income; features like occupation, age, and city, as well as their higher-order combination <occupation, age, city>, are evidently more significant than features such as weight, blood type, and zodiac sign, with their corresponding higher-order combination <weight, blood type, zodiac sign>. Therefore, the implementation of Selective Kernel Networks (SENets) could be introduced to dynamically learn the importance of different feature crosses.To dynamically learn the importance of feature interaction combinations in a fine-grained manner and to enhance the model's recommendation performance and interpretability, this paper introduces Selective Kernel Networks (SKNet) and proposes the FiiNet method (Multiple Order Feature Interaction Importance Neural Networks).

\section{our proposed model-FiiNet}

For the model's input features x, Selective Kernel Networks (SKNet) are utilized to construct multi-order feature cross combinations and further learn the importance of feature interactions dynamically in a fine-grained manner. The model is composed of the following parts: sparse input layer and embedding layer, SKNets layer, multiple hidden layers, and output layer. The sparse input layer and embedding layer are akin to those in NFM\cite{b9}. Sparse representation is applied to input features, embedding the raw feature inputs into dense vectors. The SKNet layer begins to explicitly construct multi-order feature crosses and adaptively adjusts their feature cross embedding sizes based on stimulus content, performing weighted operations on multi-order feature cross information to obtain the importance information of feature cross combinations. Subsequently, this feature cross importance information is sent to the combination layer for merging and then input into the deep neural network layers, where the network layers output prediction scores.

\subsection{ Sparse Input Layer and Embedding Layer}

Click-through rate prediction (CTR) models employing deep learning methods generally utilize sparse input layers and embedding layers, as seen in NFM\cite{b9} and FM\cite{b6}. The sparse input layer uses a sparse representation of the original input features. The embedding layer is capable of embedding the sparse features into low-dimensional, dense real-valued vectors. The output of the embedding layer is a wide concatenated domain embedding vector: $E=[e_1,e_2,\cdots,e_i,\cdots,e_f]$, where f denotes the number of feature fields in the data, $e_i\in R^K$ represents the embedding of the i-th feature field, and k denotes the dimensionality of the embedding layer.

\subsection{Selective kernel network Layer}

Usually, the influence of each feature in the target task is not the same. For example, when predicting personal income, attributes related to occupation generally have higher predictive importance than hobbies or interests. Inspired by the success of SKNet in computer vision, integrating this method allows the model to pay more attention to the impact of key features and their interactions. The SKNet method can dynamically increase the weights of key features and their interactions while compressing the proportion of those feature combinations that contribute less.

\begin{figure}[H]
	\centerline{\includegraphics[width=8cm]{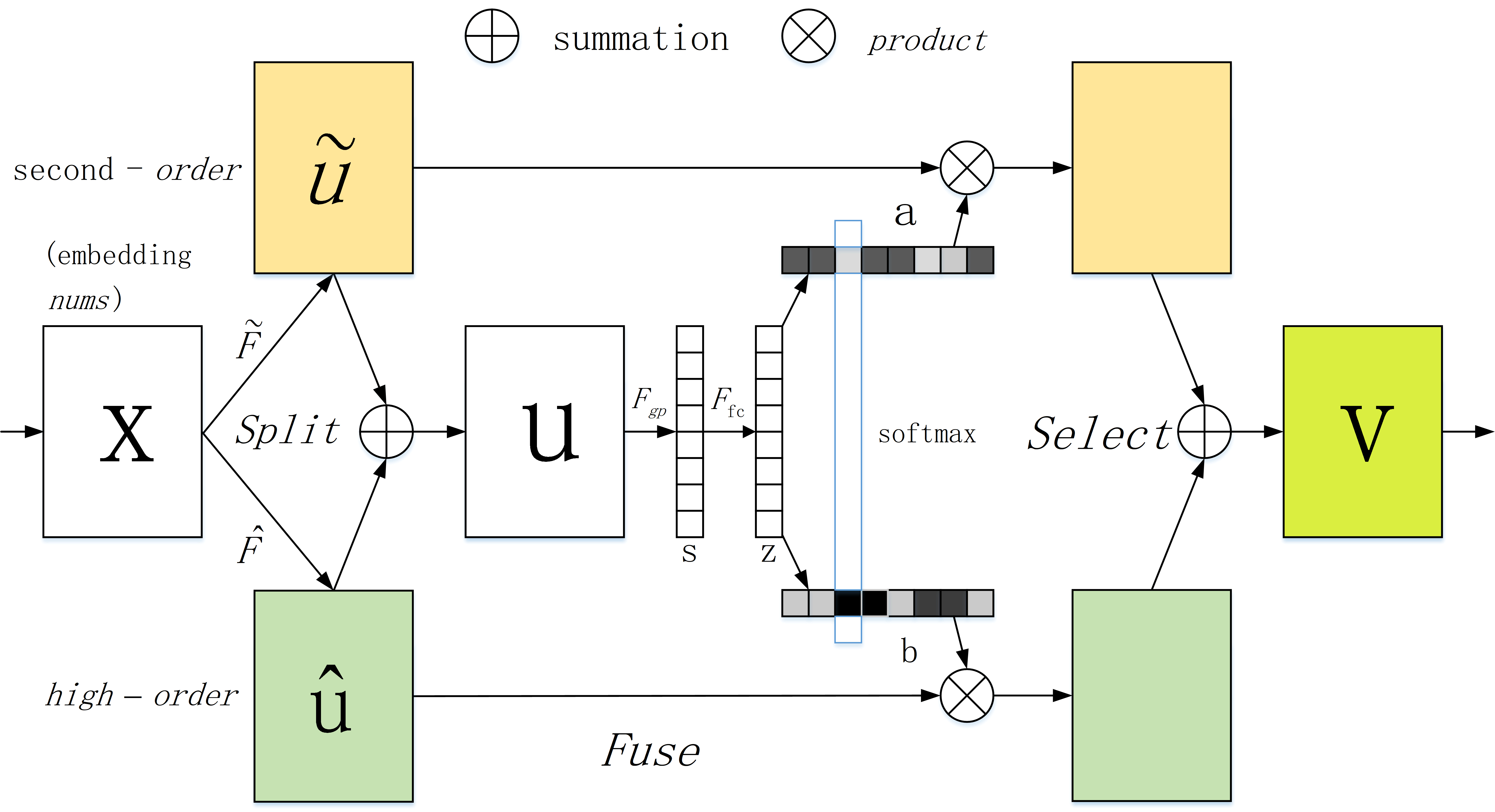}}
	\caption{SKNet mechanism layer.}
	\label{fig1}
\end{figure}

\indent To enable the neural network to adaptively learn the importance of multi-order feature crosses, selective kernel convolution is employed among multiple feature crosses with different orders. As shown in the above Figure \ref{fig1}, which is the case with two branches. Therefore, in this example, there are only two feature crosses of different orders, which can be easily expanded to cases with more branches. Specifically, the SK kernel unit operation is implemented through three operators: Split, Fuse, and Select, which can be detailed as follows:\\

\indent\textbf{1 Split}: For the feature embedding x passed from the upper layer, filters of different kernel sizes are used for sampling. For 2D feature embedding, the feature latent vectors in FM are used to construct feature crosses as shown in Equation (1):
\begin{equation}
	\sum_{i=1}^{n-1}\sum_{j=i+1}^{n}{w_{ij}x_ix_j}\label{eq}
\end{equation}
where n represents the number of features in the sample, $x_i$ is the i-th feature, and $w_{ij}$ is the learnable parameter after crossing the i-th and j-th features. This effectively constructs a second-order feature cross $\widetilde{F}$: $X\rightarrow\widetilde{U}$, and can further construct a higher-order as shown in equation (2):
\begin{equation}
	\sum_{i=1}^{\mathrm{n-2}}\sum_{j=i+1}^{n-1}\sum_{k=j+1}^{n}{w_{ijk}x_ix_jx_k}\label{eq}
\end{equation}
representing a third-order feature cross $\hat{F}$ : $X\rightarrow\hat{U}$. $\widetilde{F}$ and $\hat{F}$ consist of feature crosses of different orders. To further improve efficiency, the incoming feature sequence is fixed, allowing direct slicing of features for crossing, reducing the need for query and comparison operations after each feature sequence is passed in.\\

\indent\textbf{2 Fuse}: This operator enables the neural network to adaptively adjust the size of their feature cross embeddings based on the content of the stimulus. The basic idea is to use gates to control the flow of information carrying different scale information from multiple branches into the next layer of neurons. The information from several branches is started to be integrated by pairing, using an element-wise summation approach shown above: $U=\widetilde{U}+\hat{U}$, and then global information is embedded using global average pooling to generate channel-level statistics. Specifically, pooling methods such as max or mean are used to compress the original embedding $U=[e_1,\cdots,e_f] $ into a statistical vector $Z=[z_1\cdots,z_i\cdots,z_f]$, where $i\in[1,\cdots,f]$, and $z_i$ is a scalar value representing the global information about the i-th feature cross representation, which can be computed by the following global mean pooling equation (3):
\begin{equation}
	z_i=F_{gp}(e_i)=\frac{1}{k}\sum_{t=1}^{k}e_i^{(t)}\label{eq}
\end{equation}
In the original literature, SKNet's compression function is max pooling. However, experiments for recommendation systems show that mean pooling is more effective in capturing global information.\\
\indent Furthermore, likely SENet\cite{b15} to learn weights within the statistical vector Z, the combination of two fully connected layers (FC) is used; the first FC layer uses parameter $w_1$ for dimension reduction, with a  reducible ratio r as a tunable hyperparameter, and $\sigma_1$ as a non-linear activation function; the second FC layer employs parameter $w_2$ to increase the dimensionality. The two fully connected layers calculate the feature cross embedding weights as follows (equation 4):
\begin{equation}
	S=F_{fc}(Z)=\sigma_2(w_2\sigma_1(w_1Z))\label{eq}
\end{equation}
where S is the weight vector, $\sigma_1$ and $\sigma_2$ are activation functions, and $w_1$ and $w_2$ are learnable parameters, with r being the reduction ratio.\\

\indent\textbf{3 Select}: The Select operation corresponds to the Scale operation in the SENet module. The difference is that the Select operation uses two different scale weight matrices to weight the split multi-order feature cross information separately and then sum to obtain the final output vector V. Specifically, cross-channel soft attention is commonly used to adaptively select information from different spatial scales, which is guided here by the compact feature description factor s, applying the softmax operator across channel numbers (equation 5):
\begin{equation}
	a_c=\frac{e^{A_cs}}{e^{A_cs}+e^{B_cs}},b_c=\frac{e^{B_cs}}{e^{A_cs}+e^{B_cs}}\label{eq}
\end{equation}
where A, B and a, b represent the soft attention vectors $\widetilde{U}$ and $\hat{U}$. Note $A_c$ is the c-th row of A and $a_c$ is the c-th element of a, similarly for $B_c$ and $b_c$. In the two-branch case, since $a_c+b_c=1$, matrix B is redundant, and the final feature map V is obtained by attention weights across each kernel (equation 6):
\begin{equation}
	V_c=a_c\cdot{\widetilde{U}}_c+b_c\cdot{\hat{U}}_c,a_c+b_c=1\label{eq}
\end{equation}
where $V=[V_1,V_2,\cdots,V_C]$, c represents the total number of multi-order feature cross combinations. The explanation and equation expression for the two-branch situation is provided, and through the above expansion equation, selecttive kernel network layer can be easily extended to multi-branch situations.

\subsection{Deep Neural Network Layer}

The deep neural network consists of several fully connected layers stacked together, implicitly capturing high-order feature interactions. The input to the deep neural network layer is the output from the previous layer. Let $a^{(0)}=[c_1,c_2,\cdots,c_n] $ represent the output from the previous layer, where n is the product of the number of all multi-order feature cross combinations and the embedding dimension of the sparse feature vector. $a^{(0)}$ is fed into the deep neural network, and its feed-forward process is as follows (equation 7):
\begin{equation}
	a^{(l)}=\sigma\left(w^{\left(l\right)}a^{\left(l-1\right)}+b^{\left(l\right)}\right)\label{eq}
\end{equation}
where l denotes the depth of the network, $\sigma$ is the activation function, and $w^{(l)}$,$b^{(l)}$,$a^{(l)}$ are the model weights, bias, and output of the l-th layer, respectively. Then, a dense real-valued feature vector is generated and finally input into the sigmoid function for CTR prediction.
\subsection{Output Layer}
The overall equation for the model's output layer is (equation 8):
\begin{equation}
	\hat{y}=\sigma(w_0+\sum_{i=0}^{m}{w_ix_i+y_d})\label{eq}
\end{equation}
where $\hat{y}\in$(0,1) is the predicted CTR value, $\sigma$ is the sigmoid activation function, m denotes the feature dimension, x represents the input, and $w_i$ is the i-th dimension of the linear layer part. The learning process aims to minimize the following target function (cross-entropy) (equation 9):
\begin{equation}
	loss=-\frac{1}{N}\sum_{i=1}^{N}{(y_i\mathrm{log} (\mathrm{\ } {\hat{\mathrm{y}}}_i)+(\mathrm{1-} \mathrm{y}_i)\ast l o g{(}1-{\hat{y}}_i))}\label{eq}
\end{equation}
where $y_i$ is the true value of the i-th instance, ${\hat{y}}_i$ is the predicted CTR value of the i-th instance, and N is the total number of samples.

\section{Experiment and result analysis}

\subsection{Data sets and experiment setup}

\textbf{1 Data Sets}\\
\indent The following Table \ref{tab2}, two real-world datasets were employed for model comparative experiments, specifically including the KuaiRec big matrix subset of the KuaiRec dataset and the Book-Crossing dataset.

\begin{table}[H]
	\caption{The statistical information of the data set is used in this study}
	\begin{center}
		\begin{tabular}{c|c|c}
			\toprule 
			\textbf{Data sets}&\textbf{	KuaiRec-big}&\textbf{Book-Crossing} \\
			\midrule 
			Number of users&7176&278858 \\
			Item quantity&10728&271379 \\
			Number of visits&12530806&1149780 \\
			Average number of user accesses&1746&4 \\
			Average number of items accessed&1168&4 \\
			sparsity&16.3\%&0.0015\% \\
			\hline
			Other&Video category,&Publisher, year of \\
			attributes&user age, etc&publication, etc \\ 
			\bottomrule 
		\end{tabular}
		\label{tab2}
	\end{center}
\end{table}

\indent The Book-Crossing dataset\footnote{\url{http://www2.informatik.uni-freiburg.de/~cziegler/BX/}} is a book rating dataset that contains rating records of books by users from the Book-Crossing community, and also includes information such as users' age, location, as well as the books' publication year, publisher, and authors. The user rating for books ranges from 0 to 10, with the threshold for a user liking a book set at above 6.\\
\indent The KuaiRec-big dataset\cite{b20} is a recommender system dataset jointly released by the University of Science and Technology of China and the Kuaishou Community Science Research Laboratory. Version 2.0 of the dataset is the first to contain a dense exposure data set with interactions on the magnitude of millions. The dataset also records information such as user attributes, video attributes, and the ratio of the duration the user watches a video to the length of the video itself. This paper uses information from the KuaiRec big matrix dataset, combined with statistics on the ratio of the duration a user watches a video to the video's length, with the threshold for a user liking a video set at above 3.\\

\indent\textbf{2 Experimental Settings}\\
\indent This paper conducts experiments on two datasets, KuaiRec-big and Book-Crossing, and the following Table \ref{tab3} presents the configurations related to the experiment and the settings of model-related parameters.
\begin{table}[htbp]
	\caption{Experimental configurations and model parameter settings}
	\begin{center}
		\begin{tabular}{ccc}
			\toprule 
			\textbf{Configurations}&\textbf{Environment}&\textbf{Value} \\
			\midrule 
			&CPU&Intel Xeon(R)Gold 6430 \\
			&GPU&NVIDIA GeForce RTX4090 24GB \\
			&Memory&120GB \\
			&OS&Ubuntu 20.04 LTS \\
			&Language&Python 3.8 \\
			&Framework&Pytorch 1.11.0 \\
			\hline
			Parameter settings& & \\
			\hline
			&batch size &256 \\
			&learning rate &0.001356 \\
			&weight decay &1e-5 \\
			&epochs&500 \\
			&optimizer&adam \\
			&dropout&0.2 \\
			&embeddings&32 \\
			&initializer&Xavier \\
			&seed&2023 \\
			&loss function&bce loss \\
			\bottomrule 
		\end{tabular}
		\label{tab3}
	\end{center}
\end{table}
\subsection{Model performance analysis}

Table \ref{tab4} in this section shows the experimental results of FiiNet as a click-through rate prediction model. Its performance analysis indexes include AUC\cite{b21} (Area Under ROC) and Logloss (cross entropy loss).
\begin{table}[H]
	\caption{CTR methods are compared on test data sets}
	\begin{center}
		\begin{tabular}{c|c|c|c|c}
			\toprule 
			\textbf{CTR}&\multicolumn{2}{|c|}{\textbf{KuaiRec-big}}&\multicolumn{2}{|c}{\textbf{Book-Crossing}} \\
			\cline{2-5}
			\textbf{Model}&\textbf{{AUC}}&\textbf{{Logloss}}&\textbf{{AUC}}&\textbf{{Logloss}} \\
			\midrule 
			LR&0.7347&0.1400&0.6323&0.6605 \\
			FM\cite{b6}&0.7759&0.1353&0.6484&0.6771 \\
			AFM\cite{b8}&0.7542&0.1386&0.6392&0.6622 \\
			FFM\cite{b7}&0.7709&0.1356&0.6506&0.6613 \\
			Wide\&Deep\cite{b5}&0.7867&0.1332&0.6720&0.6387 \\
			DeepFM\cite{b10}&0.7886&0.1326&0.6603&0.6652 \\
			NFM\cite{b9}&0.7789&0.1354&0.6533&0.6514 \\
			Deep\&Cross\cite{b11}&0.7862&0.1334&0.6707&0.6410 \\
			xDeepFM\cite{b12}&0.7862&0.1333&0.6619&0.6646 \\
			FiBiNet\cite{b13}&0.7877&0.1332&0.6698&0.6403 \\
			AutoInt\cite{b14}&0.7891&0.1337&0.6534&0.6804 \\
			\hline
			FiiNet(\textbf{ours})&\textbf{0.7983 $\uparrow$}&\textbf{0.1310 $\downarrow$}&\textbf{0.6745 $\uparrow$}&\textbf{0.6369 $\downarrow$} \\
			\bottomrule 
		\end{tabular}
		\label{tab4}
	\end{center}
\end{table}
(1) Through the experiments, it was found that FiiNet's AUC indicator on the KuaiRec-big and Book-Crossing datasets was higher than that of the aforementioned click-through rate prediction models, and the Logloss indicator was lower than that of the aforementioned models. Thus, FiiNet surpassed the previous models in both performance metrics, verifying the effectiveness of Selective Kernel Networks (SKNets) in dynamically learning multi-order explicit feature crosses in a fine-grained manner and enhancing model performance. \\
\indent (2) Click-through rate prediction models incorporating high-order feature crosses outperformed those with only low-order feature crosses on both metrics, indicating that high-order feature cross combinations are also important in the model and can provide good generalization performance, thereby enhancing model performance. \\
\indent (3) The AutoInt model is a robust benchmark on the KuaiRec-big dataset, outperforming other high-order feature cross models, but its performance on the Book-Crossing dataset is rather average, similar to the low-order feature cross model FFM. This might be due to the Book-Crossing dataset being overly sparse, with the user's average visit count in the single digits and the item's average visit count also in the single digits, with data sparsity less than one ten-thousandth. Nonetheless, the FiiNet model still achieved good results.

\subsection{Embedding dimension analysis}
\begin{figure}[H]
	\centering
	\subfloat[AUC on the Book-Crossing validation set.]{\includegraphics[width=1.7in]{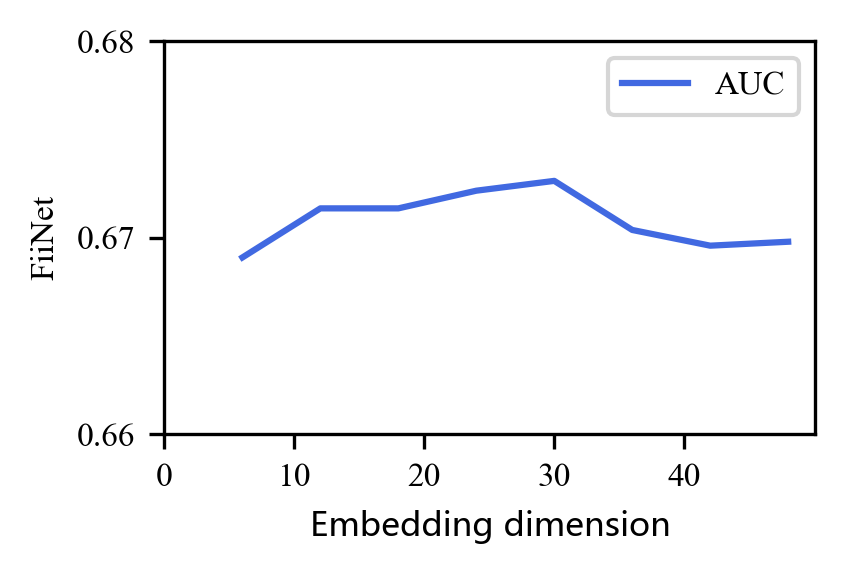}%
		\label{2.1}}
	\subfloat[AUC on the KuaiRec-big validation set.]{\includegraphics[width=1.7in]{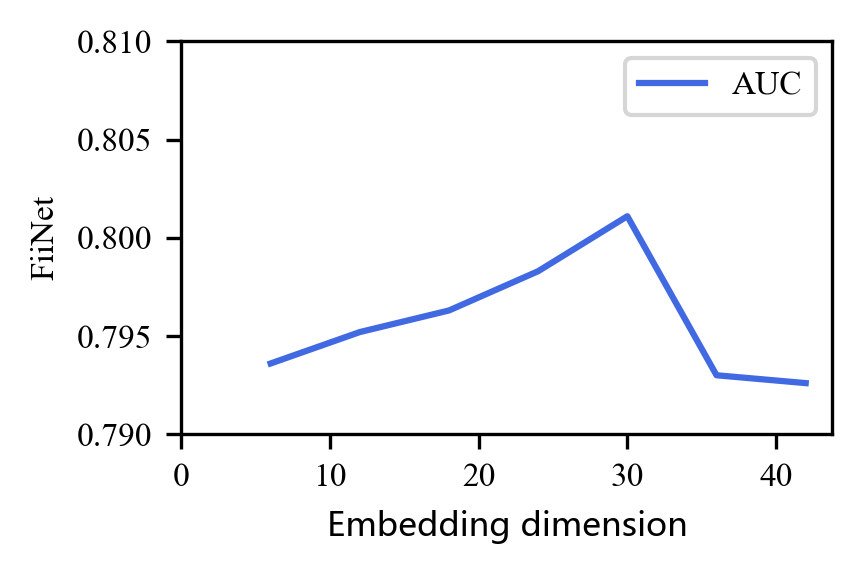}%
		\label{2.2}}
	\caption{The index on the verification set varies with the embedding dimension.}
	\label{fig2}
\end{figure}

As one of the key hyperparameters of the model, the embedded dimension has a significant impact on the performance of the model. To gain insight into this effect, the dimensions of the embedding vectors in this section are [6,12,18,24,30,36,42,48]. Figure. \ref{fig2} illustrates the sensitivity of FiiNet to embedding dimension parameters.
This study selected two different datasets, KuaiRec-big and Book-Crossing, to conduct a comprehensive analysis of the performance changes of the models at different embedding dimensions using the AUC metric:\\
\indent(1) The experimental results showed that as the embedding dimension gradually increased, the model exhibited a clear upward trend in AUC performance, indicating that the embedding dimension has a significant impact on the prediction ability of the model. Moreover, the performance of the models peaked on both datasets when the embedding dimension reached about 30. As the embedding dimension continued to increase beyond this critical point, the AUC performance of the models began to decline.\\
\indent(2) Taking into account the performance of the model and resource efficiency, the embedding dimension was ultimately fixed at 32 for practical applications. This decision is based on the model achieving good performance indicators at this dimension while maintaining a reasonable balance in computational resource and time costs.

\subsection{Ablation experiments}
\begin{figure}[H]
	\centering
	\subfloat[Comparison results of AUC indexes of four experimental models.]{\includegraphics[width=1.7in]{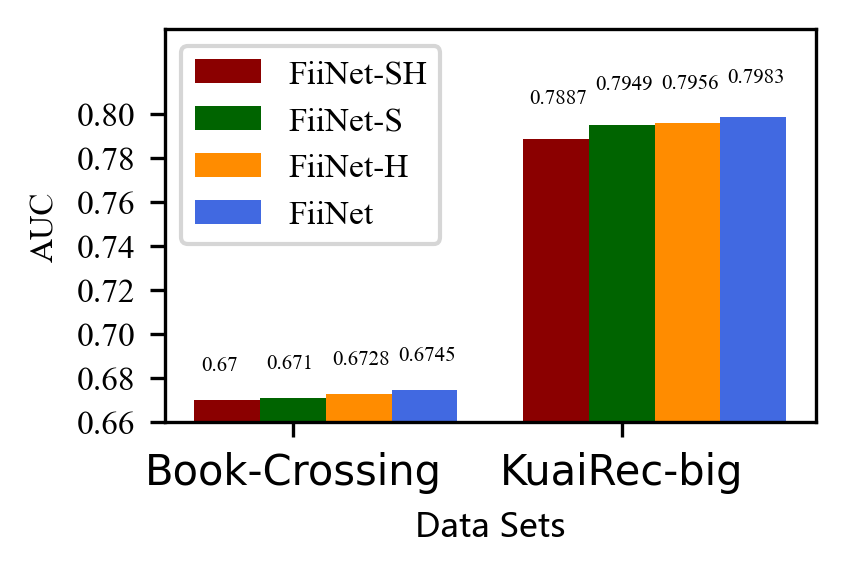}%
		\label{3.1}}
	\subfloat[Comparison results of Logloss indexes in four experimental models.]{\includegraphics[width=1.7in]{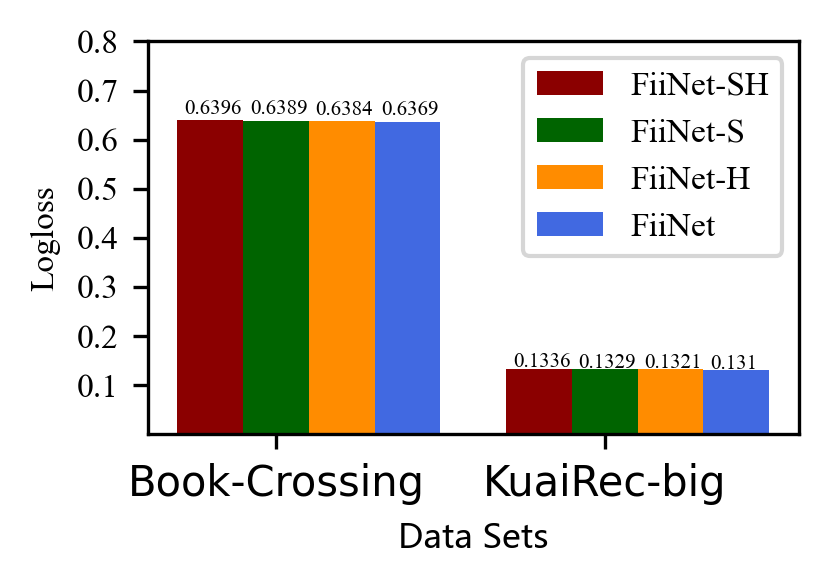}%
		\label{3.2}}
	\caption{Multi-index comparison results of four experimental models on the datasets.}
	\label{fig3}
\end{figure}

This section is devoted to conducting a series of ablation studies, aiming to explore in depth the role played by each component in the FiiNet model. By constructing several experimental variant models, the specific impact of different modules on the performance of the FiiNet model can be further analyzed and verified. In addition to the FiiNet model itself, several other experimental models are tested, including:\\
\indent(1) FiiNet-SH Model: Based on the FiiNet model, without the Selective Kernel Network's module that learns the importance of multi-order feature crosses removed, and connected to the deep network module to form the resultant model.\\
\indent(2) FiiNet-S Model: Based on the FiiNet model, without the module that learns second-order explicit feature crosses in the Selective Kernel Network removed, leading to the obtained model.\\
\indent(3) FiiNet-H Model: Based on the FiiNet model, without the module that learns higher-order explicit feature crosses in the Selective Kernel Network removed, resulting in the derived model.\\
\indent The experiments in this section are conducted on the KuaiRec-big and Book-Crossing datasets, with the aim of comprehensively evaluating the model's recommendation effects on different types of data. The detailed experimental results are all displayed in Figure. \ref{fig3}, from which the model's performance can be visually observed, and the following conclusions can be analyzed.
(1) The FiiNet model proposed in this study has demonstrated superior overall performance compared to the other three reference experimental models, as evidenced by a series of comparative experiments. Among the variants of the FiiNet model, the effects of FiiNet-S and FiiNet-H are more prominent than those of FiiNet-SH. This result robustly supports the effectiveness of the Selective Kernel Network architecture introduced in this study.\\
\indent(2) The performance of the FiiNet-H model is consistently superior to that of the FiiNet-S model. Therefore, compared to high-order explicit feature crosses in the dataset, low-order explicit feature crosses are more important for performance enhancement. One possible reason for this phenomenon might be that the sparsity degree of low-order feature cross matrices is far lower than that of high-order feature crosses, enabling the Selective Kernel Network to effectively learn the importance of the low-order feature cross matrix.\\
\indent(3) Compared to the FiiNet-SH model, the FiiNet model shows greater performance improvement on the KuaiRec-big dataset than on the Book-Crossing dataset. A possible explanation is that the Book-Crossing dataset is very sparse, with an average user access count in the single digits, and the average item access count is also in the single digits, with a sparsity degree of less than one ten-thousandth. In contrast, on the KuaiRec-big dataset, both the average user access count and the average item access count exceed one thousand, with a sparsity degree of over 15\%. The latter data set has a lower degree of sparsity in item attributes, which also verifies that the Selective Kernel Network can better learn the potential relationships between data attributes.

\subsection{Analysis of Feature Cross Importance}

\begin{figure}[H]
	\centerline{\includegraphics[width=8cm]{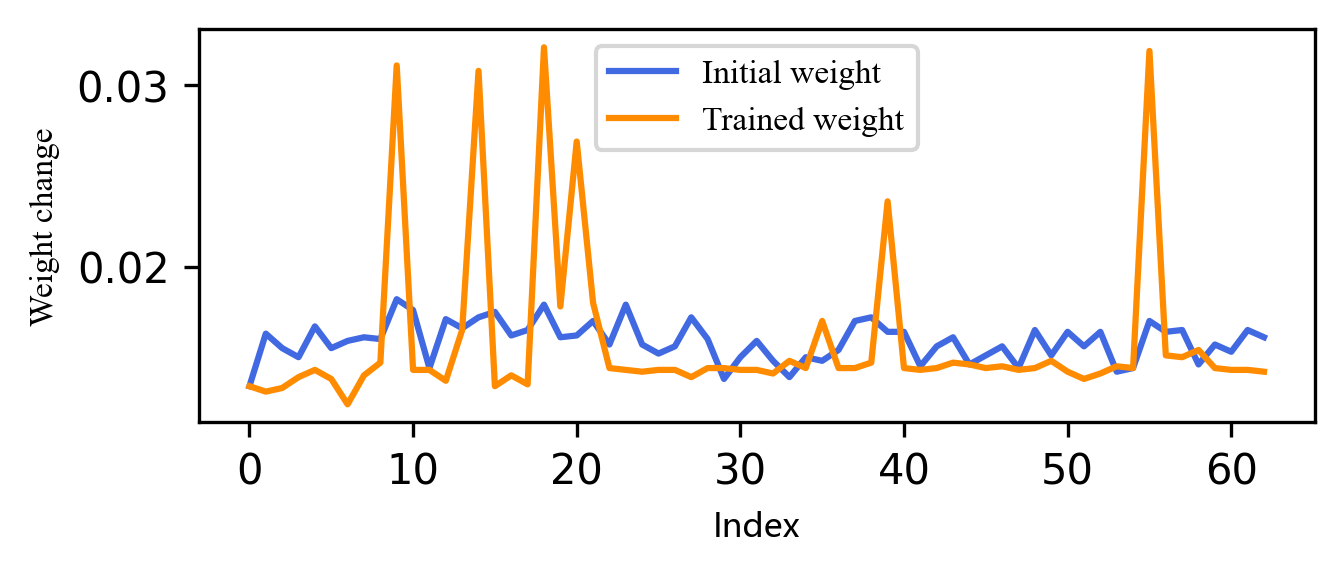}}
	\caption{Comparison of different weights of cross-combinations of multi-order features before and after training on the Book-Crossing dataset.}
	\label{fig4}
\end{figure}

The FiiNet model's recommendation system not only provides effective recommendation results but also exhibits substantial interpretability. This section details the mechanisms employed by the FiiNet model to elucidate the recommended outcomes. Using the Book-Crossing dataset as an instance, one can examine the recommendations suggested by the FiiNet algorithm, specifically items favored by users. Figure. \ref{fig4} illustrates a comparison between the weights of different parameters before and after training multi-order feature interactions, with these parameter weights being derived by procuring the model's attention weights.
(1) By comparing the attention weights of different feature crosses before and after training, it is evident that the trained FiiNet is able to discern meaningful feature cross combinations, increasing the attention weights for important feature crosses, while diminishing those for non-informative feature crosses.\\

\indent(2) Moreover, within the SKNets layer of the FiiNet model, both low-order and high-order feature cross combinations are constructed. In Figure. \ref{fig4}, the earlier section of the index coordinates corresponds to low-order feature crosses, while the latter part pertains to high-order feature crosses. It is observable that there are more combinations with higher attention weights in the low-order feature crosses, and fewer in the high-order feature crosses, indicating that the low-order feature crosses have a more significant impact on the model's performance. This observation aligns with the conclusion 2 of the ablation experiments section.

\section{Conclusions}
This paper incorporate a selective kernel network based on the channel attention mechanism for personalized recommendation, and design FiiNet, a click-through rate prediction method that uses a selective kernel network to dynamically learn the importance of the intersection of multi-order features. It confirms that the FiiNet model can increase the attention weights of critical feature cross combinations while reducing those of non-informative feature crosses during training, thereby enhancing the model's recommendation performance and interpretability by effectively utilizing combinations of feature crosses. Subsequently, the proposed model was validated through experimental results on two public real-world datasets. In the experimental analysis, ablation experiments were conducted to demonstrate the effectiveness of each component of the model; in the section analyzing the importance of feature cross combinations, the significance of different feature crosses was confirmed, corroborating the conclusions in the ablation experiments and further providing interpretability to the model. The experiments showed that the FiiNet model outperforms other click-through rate prediction models.


\addtolength{\textheight}{-12cm}   






\bibliographystyle{ieeetr}
\bibliography{references.bib}

\end{document}